# Convolutional Neural Networks for Automatic Detection of Intact Adenovirus from TEM Imaging with Debris, Broken and Artefacts Particles

Olivier Rukundo[a,b], Andrea Behanova[a,c], Riccardo De Feo[a,d], Seppo Rönkkö[e], Joni Oja[e], Jussi Tohka[a]

[a] A.I. Virtanen Institute for Molecular Sciences, University of Eastern Finland, Kuopio, Finland
[b] Center for Clinical Research, University Clinic of Dentistry, Medical University of Vienna, Vienna, Austria
[c] Department of Information Technology, Uppsala University, Uppsala, Sweden
[d] Charles River Laboratory, Kuopio, Finland
[e] FinVector Oy, Kuopio, Finland

Corresponding author: olivier.rukundo@meduniwien.ac.at

**Abstract**

Regular monitoring of the primary particles and purity profiles of a drug product during development and manufacturing processes is essential for manufacturers to avoid product variability and contamination. Transmission electron microscopy (TEM) imaging helps manufacturers predict how changes affect particle characteristics and purity for virus-based gene therapy vector products and intermediates. Since intact particles can characterize efficacious products, it is beneficial to automate the detection of intact adenovirus against a non-intact-viral background mixed with debris, broken, and artefact particles. In the presence of such particles, detecting intact adenoviruses becomes more challenging. To overcome the challenge, due to such a presence, we developed a software tool for semi-automatic annotation and segmentation of adenoviruses and a software tool for automatic segmentation and detection of intact adenoviruses in TEM imaging systems. The developed semi-automatic tool exploited conventional image analysis techniques while the automatic tool was built based on convolutional neural networks and image analysis techniques. Our quantitative and qualitative evaluations showed outstanding true positive detection rates compared to false positive and negative rates where adenoviruses were nicely detected without mistaking them for real debris, broken adenoviruses, and/or staining artefacts.

1. **Introduction**

Large-scale production of viral vectors for gene therapy requires tools to characterize the virus particles [2]. Transmission electron microscopy (TEM) is the only imaging technique allowing the direct visualization of viruses, due to its nanometer-scale resolution [21], [4]. Consequently, with TEM, it becomes possible to understand what occurs with viral particles when parameters or process operations change or when formulations are modified. Different biomanufacturing process conditions have different effects on particle characteristics, and images that reveal particle morphology together with quantitative analysis can provide a good understanding of and insights into the impact of such process changes via assessing overall morphology (stability, purity, integrity, and clustering) which might affect vector performance [1], [3]. However, due to the need for considerable operator skills, special laboratory facilities, and the limitations in providing quantitative data, it is not routinely used in process development [25]. It is important to note that TEM image analysis is typically performed in specialized TEM facilities, and the time to get results is often long [25]. Also, the process to annotate, segment, and detect intact adenoviruses in TEM images remains challenging due to the presence of broken adenoviruses, debris, and various kinds of staining artefacts as illustrated in Figure 1. Consequently, the intact adenovirus segmentation in TEM images using traditional image analysis methods is not reliable [5]; challenging intact adenovirus characterization. Deep convolutional neural networks (CNNs) have shown excellent performance in many biomedical imaging tasks which were thought to be unsolvable before the deep learning era [22], [23],[24]. Here, the CNN of interest was U-net, which is widely used and known for its excellent segmentation precision of medical images [7], [16], [18]. U-Net is a modified and/or extended version of a fully convolutional network that works with very few training images to yield more precise segmentations [7], [16], [18]. Although many works currently exist, mostly proposed for segmentation of bio/medical images using the U-net or variants or closely related versions [7], [8], [9], [10], [34], the U-net outperformed the earlier best methods and could still provide a fast and accurate segmentation of images.

However, research in the automatic segmentation of intact adenoviruses in TEM images remains in its infancy. There exist a few works that proposed both CNN-based and non-CNN-based solutions to image analysis of TEM images of virus particles [11], [12], [13], [14], [15]. References [11] and [15] propose methods for segmentation of different types of viruses, including adenoviruses, from TEM images using a morphological image analysis pipeline [15] and U-Net [11]. Reference [12] proposes a method for classification between different types of viruses and makes available an open TEM dataset to study virus-type classification. Reference [13] proposes a fully connected neural network to detect feline calicivirus particles from TEM images. Finally, reference [14] focuses on the reduction of the number of trainable U-Net weights for segmentation of various virus particles from TEM images.

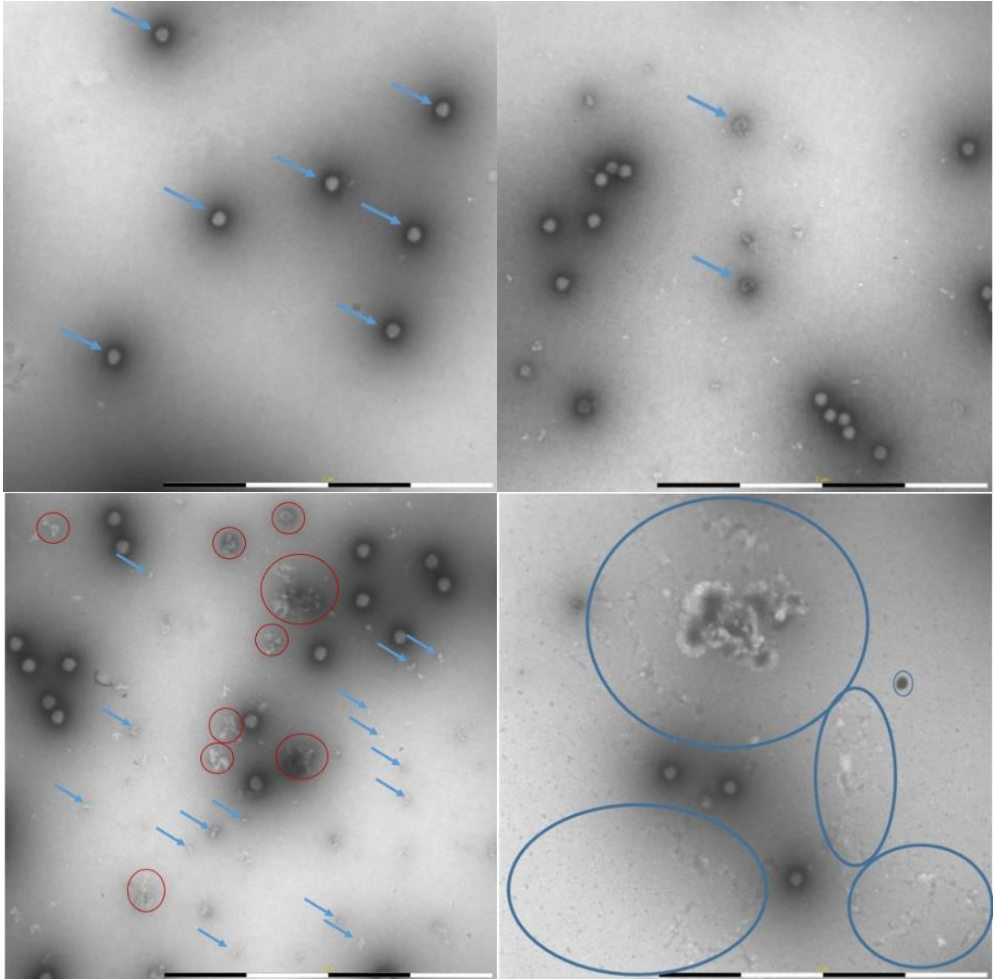

Figure 1: Intact adenovirus particles (top-left-side – blue arrow), broken adenovirus particles (top-right-side – blue arrows). debris particles (bottom-left-side: inside red circles – large debris and blue arrows – small debris), artefact particles (bottom-right-side: inside blue circles - examples of uranyl acetate staining artefacts).

However, among these works, there was no clear focus or dedicated work on intact adenovirus segmentation and detection with the aim of improving the characterization of adenoviruses in images captured by high-throughput TEM systems for production of viral vectors. Therefore, we introduce a U-Net-based approach together with software tools for fast and easy training, for segmentation of intact adenoviruses from high-throughput TEM images. Our purpose is not only due to the need for testing the automation of detection of intact adenovirus from TEM imaging with debris, broken, and artefacts particles but also to demonstrate that, detecting intact adenoviruses with high accuracy, even in highly challenging imaging conditions, was possible with U-Net.

## 2. Material and Methods

*2.1 Image data*

The imaging was performed by using the MiniTEM microscope by Vironova AB, Stockholm, Sweden [6], with an operating voltage of 25 kV and with a field of view (FOV) of 3 µm for the adenovirus samples [27]. We first acquired a training and validation set of 50 images of the size of 2048-by-2048. The intact adenoviruses of this set were annotated using a semiautomatic software tool developed by us specifically for this purpose. We used this image set to train the CNN and validate its performance using cross-validation. Second, we acquired a test set of 20 MiniTEM images that were completely independent from the training and validation set and used to test the final CNN model for adenovirus detection. This test set contained very challenging images with varying levels of debris and staining artefacts that would be too challenging for the traditional image analysis methods.

*2.2 Software tool for semi-automatic annotation and segmentation of intact adenovirus*

*2.2.1  Semi-automatic annotation*

The image annotation process is one of the most challenging steps that affect the training outcome for the automatic segmentation of microscopy images [11]. Also, annotating large enough training sets for supervised learning is a bottleneck, and dedicated tools to speed up the annotation process are still needed [28], [29], [30].

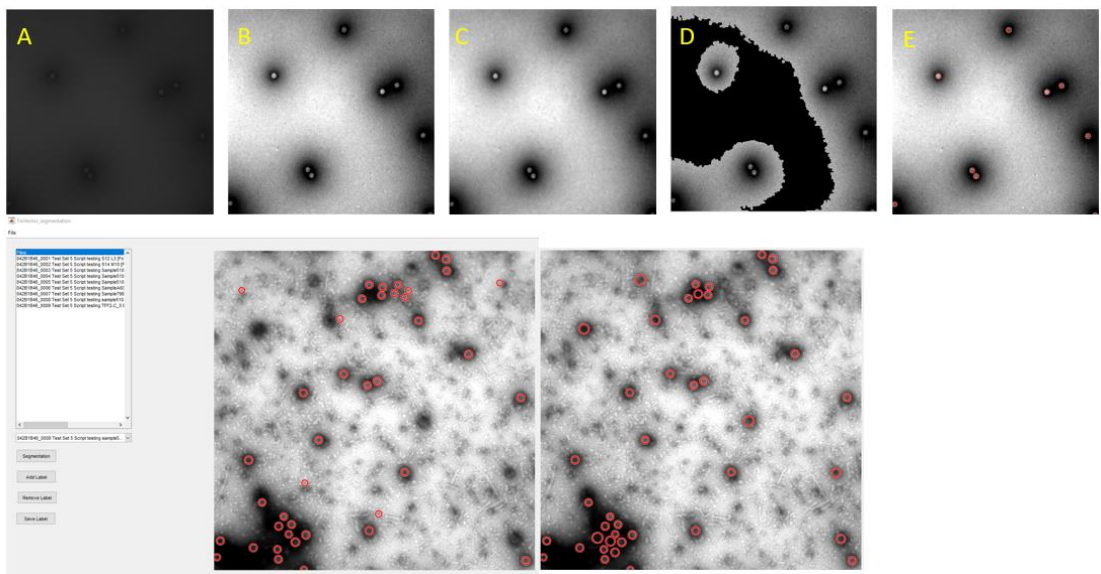

Figure 2. The developed software tool for semi-automatic segmentation of adenoviruses in MiniTEM images. Top row (A) Close-up of the original MiniTEM image, (B) contrast-enhanced image, (C) median filtered contrast-enhanced image (D) image with large bright areas masked out (E) Adenoviruses detected by Hough transform. Bottom row: Left GUI of the annotation tool with an image with the overlaid automatic segmentation, right: Image with overlaid segmentation after manual corrections.

In this regard, a GUI-based software tool for semi-automated segmentation of MiniTEM images was developed and later used to create annotated MiniTEM images used for training the U-Net model. The software tool is available at https://github.com/AndreaBehan/miniTEM-Image-Segmentation. A video showcasing the annotation process is available in the supplement. The tool can be used for rapid manual and semi-automatic annotation and semi-automatic segmentation of intact adenoviruses and other types of debris.

*2.2.2  Semi-automatic segmentation*

Using the developed semi-automatic tool required to first create a set of candidate adenoviruses through automatic image analysis operations. It is important to note that the entire procedure is based on an assumption that an intact adenovirus is a circular, bright object surrounded by a darker area. The key steps are as follows, see the panels A, B, C, D, and E of Figure 2:

1) Enhance the contrast of the image by saturating the top and the bottom 1% of intensity values in the images and perform the median filtering, with a 15 by 15 window, on the enhanced image. (Figure 2, panels B and C)
2) Segment out large bright areas of the median filtered image by thresholding followed by morphological operations. This operation is necessary to allow for the Hough transform in the next step to concentrate on adenoviruses. Note that this step does not remove intact adenoviruses as they are surrounded by a darker area (Figure 2 panel D).
3) Find adenovirus boundaries by using the circular Hough transform [31] (Figure 2 panel E)
4) Remove candidate adenoviruses that do not have a dark area surrounding them by detecting the mode of the histogram of the rectangular patch around the adenovirus.

After that, the user can interactively add and remove adenoviruses as shown in the supplementary video.

*2.3 Software tool for automatic segmentation and detection of intact adenovirus*

### 2.3.1 U-Net architecture

Our CNN for automatic segmentation was based on the U-Net architecture [16] as implemented in MATLAB. U-Net features a U-shaped design, comprising contracting and expansive paths.

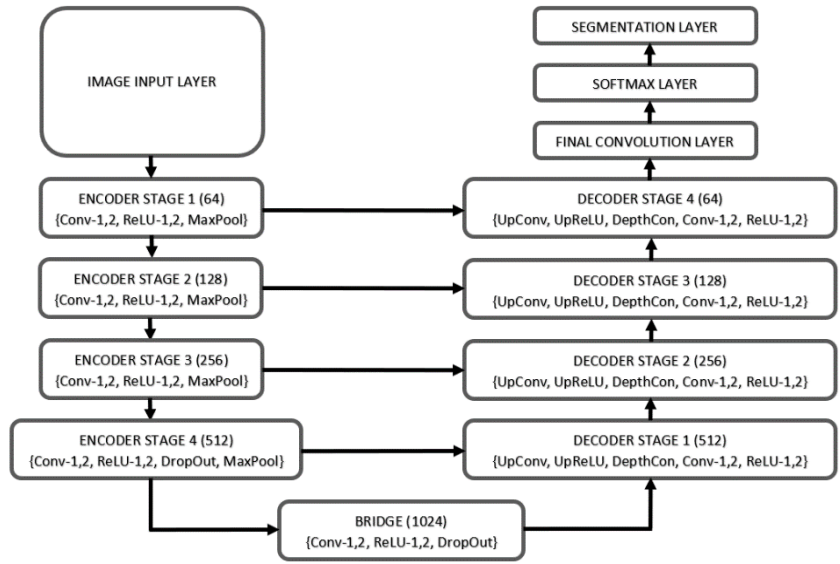

Figure 3: The U-net architecture used in this work. Conv means convolution. ReLU is a rectified linear unit. DepthConv is depth concatenation. UpConv means up-convolution or transposed convolution. MaxPool is Max Pooling

Figure 3 shows the input and output layers, as well as the intermediate layers and connections, of a deep learning network as visualized by the analyzeNetwork function in MATLAB. The contracting path consists of repeating blocks of convolution, ReLU activation, and max pooling. The expansive path involves transposed convolution, ReLU activation, concatenation with the downsampled feature map, and additional convolution.

### 2.3.2 Training

To avoid high computation demands, during the U-net training process, each 2048-by-2048 image was split into non-overlapping 64 image patches of the size 256-by-256. For each original MiniTEM image with 16 bits' depth was changed to 8 bits' depth image to minimize memory usage during the training and evaluation. The execution environment was single-GPU with the Nvidia GeForce RTX 3070 graphic card and 11[th] Gen Intel(R) Core(TM) i7-11700F @ 2.50GHz, 2496 Mhz, 8 Core(s), 16 Logical Processor(s).

### 2.3.3 Hyperparameter settings

Hyperparameter settings were manually adjusted with no new adjustments if 90% of training accuracy was reached during the first 10% of all epochs [18]. Training hyperparameters that were not listed below remained set to default, including the number of first encoder filters and encoder depth. The number of epochs = 30; the minimum batch size = 4; the initial learning rate = 0.0001; L2 regularization = 0.00005; optimizer = Adam (adaptive moment estimation algorithm). The loss function used was the default cross-entropy function provided by MATLAB's U-Net Layers function, for image segmentation using the U-Net architecture [33]. In other words, the pixel classification layer was not replaced with a weighted pixel classification layer.

### 2.3.4 Data augmentation

The data augmentation options used consisted of the random reflection in the left-right direction as well as the range of vertical and horizontal translations, with 50% probability, on the pixel interval ranging from -10 to 10.

### 2.4 Post-processing

A systematic combination of image filtering, dilating, and burning functions, [35], [36], [37], was used/applied to improve the quality of outlines of U-net's segmentation masks. In this way, we could emphasize or highlight the most precise outlines of intact adenovirus.

### 2.5 Performance evaluation metrics

We evaluated the segmentation both in terms of detection and segmentation performance. For detection, we counted the number of true positives (TP: intact adenovirus correctly detected by U-Net), false positives (FP: adenovirus incorrectly detected by the U-Net), and false negatives (FN: intact adenovirus not detected by U-Net) [19]. Based on TP, FP, and FN counts, we computed *recall* and *precision* and *F-value* as

$$precision = \frac{TP}{TP+FP} \quad (1)$$

$$recall = \frac{TP}{TP+FN} \quad (2)$$

$$F - value = \frac{(1+\beta^2)*recall*precision}{\beta^2*recall+precision} \quad (3)$$

In Eq.3, $\beta$ corresponds to the relative importance of precision versus recall, which we set $\beta = 1$ [19] [38]. We defined correct (and incorrect) detections based on the overlap of the segmentation masks and ground-truth masks, which required setting a threshold value on the overlap. To demonstrate that the detection results were not dependent on a single threshold value, we set our main or primary threshold at 75%. We also examined the secondary thresholds of 50% and 25%, as illustrated in Figure 4. For the external test set, for which manually created ground-truth segmentations did not exist, we used the developed software tool to detect and count the number of detected and missed adenoviruses as well as those incorrectly highlighted as detected or not detected. Also, we defined the threshold at 75%, 50%, and 25%, and defining a match was subjective (see Figure 4-b).

For the evaluation of the semantic segmentation, we used the Dice score and intersection over the union (IoU) also known as the Jaccard coefficient as the performance measures [19].

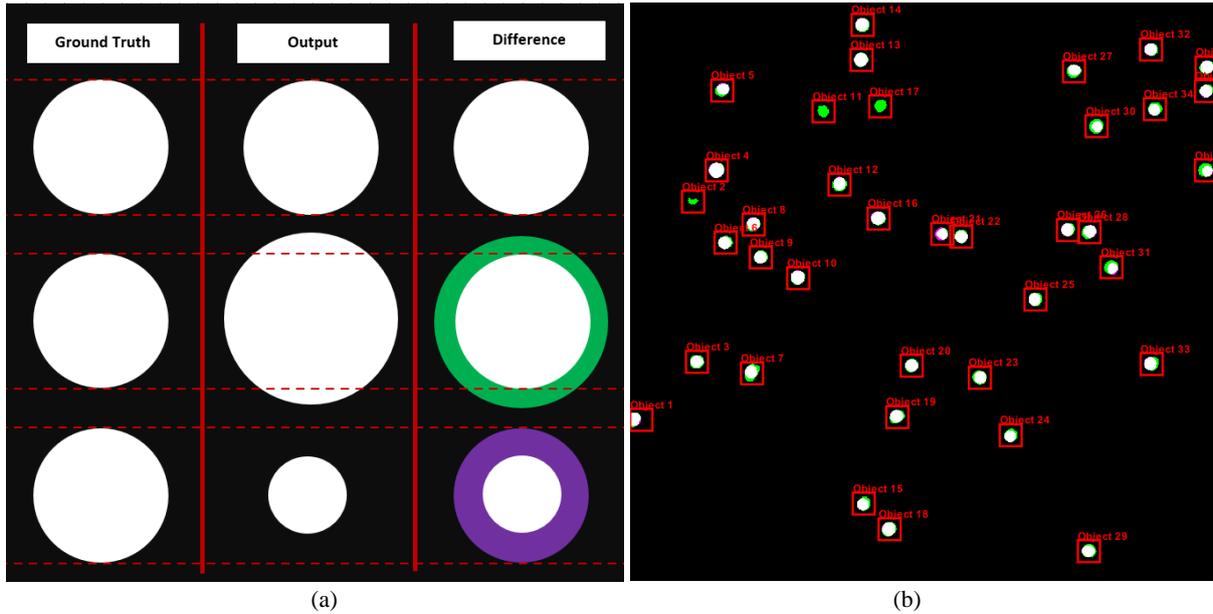

(a)  (b)

Figure 4: Example of illustration showing the ideal (a) and real (b) cases of TP, FP, and FN: In our experiments, we divided the TP situations into categories based on subjectively defined thresholds. The primary threshold was set at 75% of the area of the full circle, while secondary thresholds were set at 50% and 25% of the area of the full circle to represent the extent of differences between the ground truth and output masks. FP and FN were defined as cases where there were complete and noticeable differences between the areas of ground truth and output masks, as shown in green and purple colors, respectively.

## 3. Results

3.1 Quantitative evaluation with K-fold cross-validation on the training and validation sets

We used 5-fold cross-validation on the training and validation set to quantitatively evaluate the segmentation results. Figure 5 represents the quantitative evaluation results on detection. As the figure illustrates, on average the detection rates were high, with both the average precision and recall exceeding 90% with all the studied thresholds. On some folds, where the total number of intact adenoviruses was low, the precision and/or recall dropped below the 90% limit. However, even in these folds, the precision and recall exceeded 80% in most cases indicating sufficient precision and recall for practical applications to monitor the manufacturing process of a drug product. For the segmentation evaluation, the average Dice score exceeded 0.80, which indicates that the segmentation quality corresponded well to the ground truth. In fact, perfect segmentation results were not expected here due to the semi-automated nature of the annotation of the training data and the potential difficulty in setting the boundary of the intact adenovirus. The segmentation quality was more than sufficient to assess the morphology of intact adenoviruses in monitoring the manufacturing process.

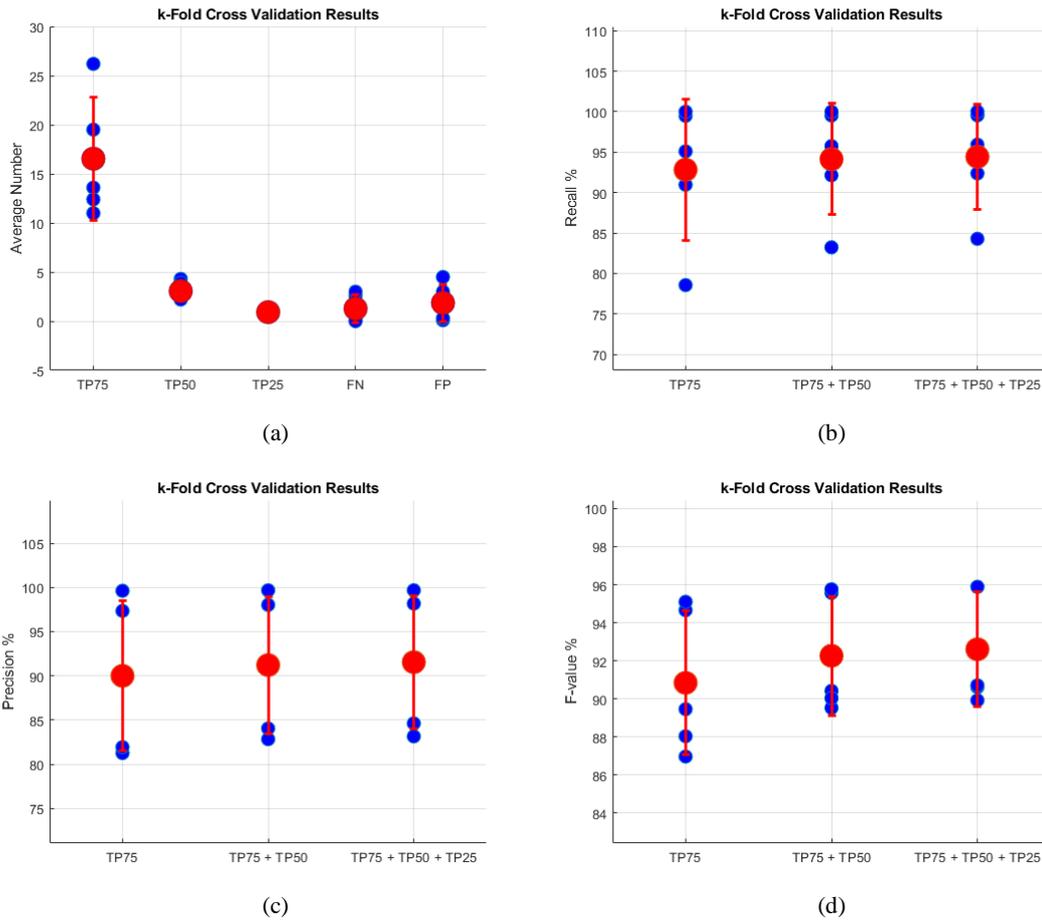

Figure 5: (a) Average number, (b) Recall, (c) Precision, and (d) F-value. Blue dots represent the results corresponding to the individual cross-validation folds, and the red dot is their average. TP stands for true positive. TP75 represents our main threshold set at 75%, TP50 represents the secondary TP threshold set at 50%, and TP25 is another secondary TP threshold set at 25%. TP75 + TP50 refers to the case, where we count the detections with more than 50% overlap with the ground-truth segmentations as correct. TP75 + TP50 + TP25 refers to the case, where we count the detections with more than 25% overlap with the ground-truth segmentations as correct.

Figure 6 presents the average percentage of intact adenovirus detected in terms of Dice and Intersection over Union (IoU) scores for each of the 5-folds.

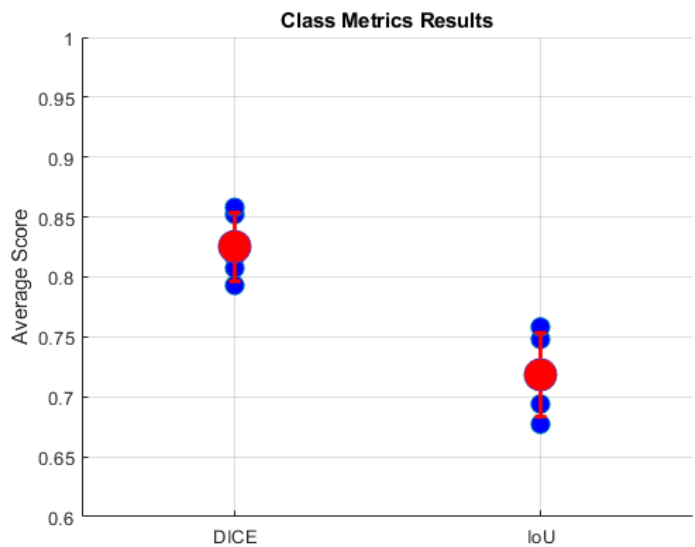

Figure 6: Average Dice and IoU score. Blue dots represent the results corresponding to the individual cross-validation folds, and the red dot is their average.

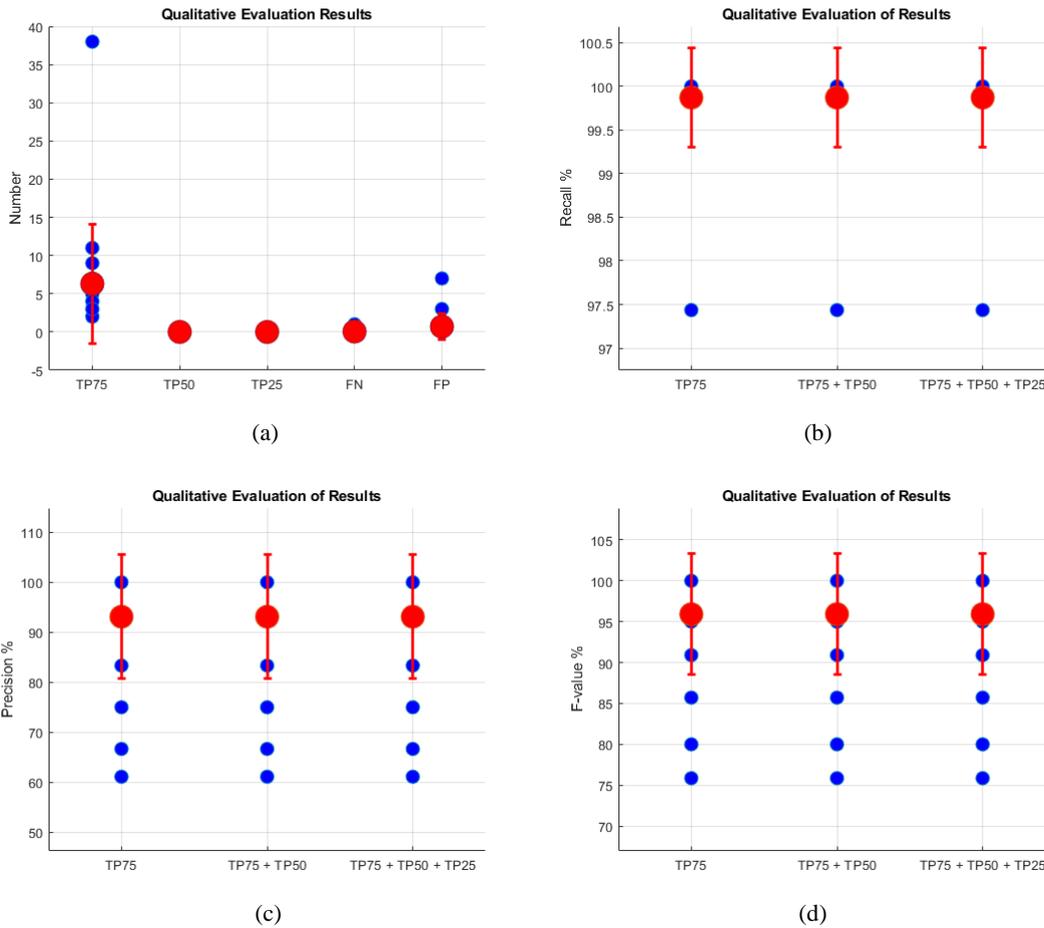

Figure 7: (a) Number, (b) Recall, (c) Precision, and (d) F-value. TP75 is our main threshold set at 75%, and TP50 and TP25 are secondary TP threshold set at 50%, and 25%, respectively. TP75 + TP50 refers to the case, where we count the detections with more than 50% overlap with the ground-truth segmentations as correct. TP75 + TP50 + TP25 refers to the case, where we count the detections with more than 25% overlap with the ground-truth segmentations as correct.

### 3.2 Results on the external test set

The external test set of 20 MiniTEM images was selected to test the accuracy of automatic detection of intact adenoviruses. These images were mixed–quality images containing intact adenoviruses, debris, artefacts, and broken particles. The quantitative detection results are shown in Figure 7 and all the 20 segmentations are shown in Figure 8. Note that we manually scored the detections as no ground-truth segmentation was available. Here, a high recall was achieved in all the cases, but the precision, albeit higher than 90% on average, remained low in some of the cases. Figure 8 shows the segmentation overlaid on the images, suggesting good segmentation performance even with the images containing non-intact adenoviruses, debris, and various staining artefacts.

### 4. Discussion

In this work, we introduced a U-Net-based system for segmentation of intact adenoviruses from high-throughput TEM images for characterization of virus particles required in the production of virus vectors. Our experimental results demonstrated a great potential for precise automated detection of intact adenovirus in TEM system images with varying degrees of quality. More interestingly, the developed software tool for automatic detection did not mistake intact adenovirus for structures or debris or artefacts similar to an internally stained particle, doublet conformation, and triplet conformation of adenoviruses. Also, it did not mistake intact adenovirus for gradually degenerated integrity adenoviruses or black spots. However, due to the presence of debris and artefacts in MiniTEM images, there were a few cases of false negative and false positive detections as shown in Section 3.

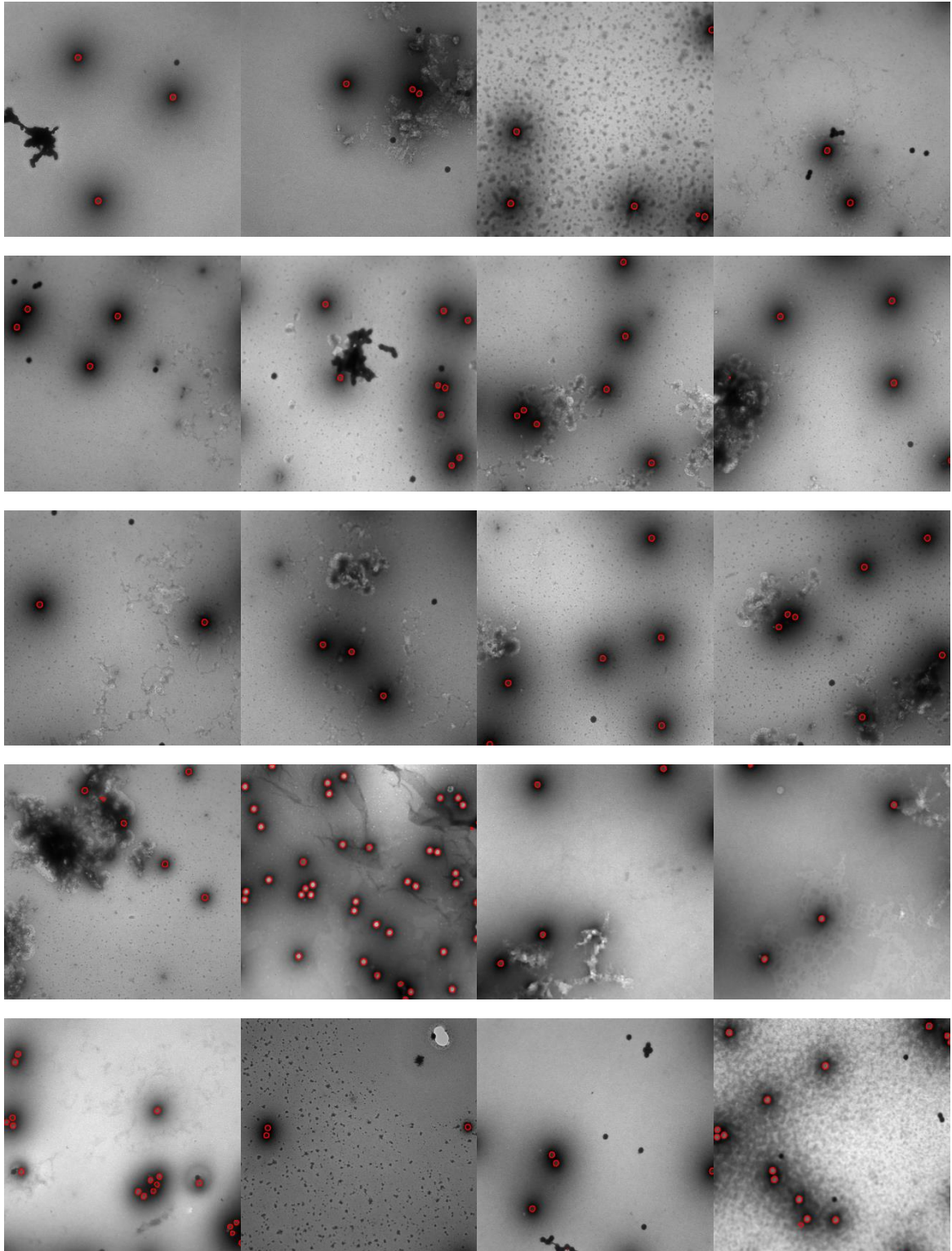

Figure 8: The automatic segmentations on the images of the external test set images.

We successfully demonstrated that it is possible to develop an automated segmentation tool for high-throughput experiments with relatively little operator effort by combining a semi-automated custom-made software tool for training and U-Net. While neural networks have been increasingly used for segmentation, detection, and classification of viruses and other particles from TEM images, there have been no previous works specifically focusing on the high-throughput imaging required in the production of virus vectors. Most of the previous work has concentrated on the detection of particles (such as human cytomegalovirus [39,40] or caveolae [41]) or the classification of different types of viruses in TEM images [12]. In the field of materials science, there is a great interest in using neural network-based approaches for characterizing nanoparticles based on TEM images [42,43,44]. Furthermore, [23] proposed the use of a fully residual U-Net for the segmentation of small extracellular vesicles from TEM images. Apart from the application itself, these image analysis problems differ considerably from the one we were facing. In our case, the main challenge lies in the variable quality of the images and the variable appearance of adenoviruses in the images acquired under different biomanufacturing process conditions, rather than the variable shape or form of adenoviruses, which are well-defined.

## 5. Conclusions

To ease the development of improving adenovirus characterization, we developed a software tool for automatic segmentation and detection of intact adenovirus in TEM imaging systems, particularly MiniTEM. Despite the presence of debris and artefacts as well as broken particles in MiniTEM images, the developed software tool demonstrated the possibility to accurately and automatically segment and detect intact adenovirus particles. Future potential research efforts may cover small, large, and rod debris definitions for automatic segmentation and quantification purposes.

**Funding**

This work was funded via the project titled "Enhancing the Innovation Potential by advancing the know-how on biomedical image analysis" by the European Social Fund (S21770).

**Author contributions**

Olivier Rukundo: Designed the convolutional neural network for automatic segmentation of adenoviruses, developed the software tool for automatic detection of intact adenoviruses, and wrote the paper. Andrea Behanova and Riccardo De Feo: Designed and developed the software tool for semi-automatic segmentation of adenoviruses and debris. Seppo Rönkkö and Joni Oja: Provided the training and testing images reviewed the paper and confirmed the validity of the study. Jussi Tohka: Read the paper and suggested modifications, conceptualized and supervised the research, and acquired the funding for the project that supported this work/paper.

**Conflict of interest**

The authors declare no conflict of interest.

**Supplementary material**

Software for semi-automatic annotation: https://blogs.uef.fi/kubiac/software/
Software for automatic segmentation: https://blogs.uef.fi/kubiac/software/
Software overview: https://www.youtube.com/watch?v=4UZJHDPKI-g